\documentclass[10pt,twocolumn,letterpaper]{article}

\usepackage[pagenumbers]{cvpr} 

\definecolor{cvprblue}{rgb}{0.21,0.49,0.74}
\usepackage[pagebackref,breaklinks,colorlinks,allcolors=cvprblue]{hyperref}

\usepackage{amsmath,amsfonts,bm}

\def\eqref#1{equation~\ref{#1}}

\def\1{\bm{1}}

\def\rvx{{\mathbf{x}}}

\DeclareMathAlphabet{\mathsfit}{\encodingdefault}{\sfdefault}{m}{sl}
\SetMathAlphabet{\mathsfit}{bold}{\encodingdefault}{\sfdefault}{bx}{n}

\newcommand{\Ls}{\mathcal{L}}

\DeclareMathOperator*{\argmax}{arg\,max}
\DeclareMathOperator*{\argmin}{arg\,min}

\usepackage{algorithm}
\usepackage{algorithmicx}
\usepackage{algpseudocode}
\usepackage{multirow}
\usepackage{makecell}
\definecolor{deepgreen}{RGB}{50, 186, 70}
\definecolor{lightgray}{rgb}{0.9, 0.9, 0.9}

\def\paperID{***} 
\def\confName{CVPR}
\def\confYear{2025}

\title{TAPT: Test-Time Adversarial Prompt Tuning for Robust Inference in Vision-Language Models}

\author{Xin Wang$^{1}$,~Kai Chen$^{1}$,~Jiaming Zhang$^{2}$,~Jingjing Chen$^{1}$,~Xingjun Ma$^{1}$\thanks{Corresponding author.} \\[0.5em]
$^{1}$Shanghai Key Lab of Intell. Info. Processing, School of CS, Fudan University \\
$^{2}$Hong Kong University of Science and Technology \\}

\begin{document}
\maketitle
\begin{abstract}
Large pre-trained Vision-Language Models (VLMs) such as CLIP have demonstrated excellent zero-shot generalizability across various downstream tasks. However, recent studies have shown that the inference performance of CLIP can be greatly degraded by small adversarial perturbations, especially its visual modality, posing significant safety threats. To mitigate this vulnerability, in this paper, we propose a novel defense method called \textbf{Test-Time Adversarial Prompt Tuning (TAPT)} to enhance the inference robustness of CLIP against visual adversarial attacks. TAPT is a test-time defense method that learns defensive bimodal (textual and visual) prompts to robustify the inference process of CLIP. Specifically, it is an unsupervised method that optimizes the defensive prompts for each test sample by minimizing a multi-view entropy and aligning adversarial-clean distributions. We evaluate the effectiveness of TAPT on 11 benchmark datasets, including ImageNet and 10 other zero-shot datasets, demonstrating that it enhances the zero-shot adversarial robustness of the original CLIP by at least 48.9\% against AutoAttack (AA), while largely maintaining performance on clean examples. Moreover, TAPT outperforms existing adversarial prompt tuning methods across various backbones, achieving an average robustness improvement of at least 36.6\%.

\end{abstract}    
\section{Introduction}
\label{sec:intro}

Vision-Language Models (VLMs) pre-trained on large-scale datasets of image-text pairs have emerged as powerful backbones for numerous applications, including computer vision~\cite{radford2021learning, jia2021scaling, zhang2023multi}, medical image analysis~\cite{huang2023visual, wang2022medclip}, and robotics~\cite{ahn2022can, shridhar2022cliport, khandelwal2022simple}. Despite these advancements, studies indicate that even minor adversarial perturbations in input images can significantly degrade the inference performance of VLMs~\cite{szegedy2013intriguing, madry2017towards, dong2018boosting, zhang2022towards, zhao2024evaluating}, thereby posing critical safety risks across a wide range of downstream applications.

Adversarial training~\cite{madry2017towards, zhang2019theoretically, gan2020large, wang2024revisiting} is a general defense strategy that augments training data with adversarial examples crafted to mislead the model. While this method has proven effective~\cite{croce2020reliable}, it entails costly min-max training, which restricts its practicality—especially for large VLMs, where standard training alone can cost millions of dollars~\cite{sharir2020cost}. To address the efficiency limitations of adversarial training, recent research has introduced adversarial prompt tuning (APT) methods~\cite{zhang2023adversarial, li2024one, zhou2024few}, which align learnable text prompts with adversarial image embeddings. These differentiable prompts~\cite{zhou2022learning, zhou2022conditional, khattak2023maple} provide a more cost-effective alternative to adversarial training.

\begin{figure}
    \centering
    \includegraphics[width=1\linewidth]{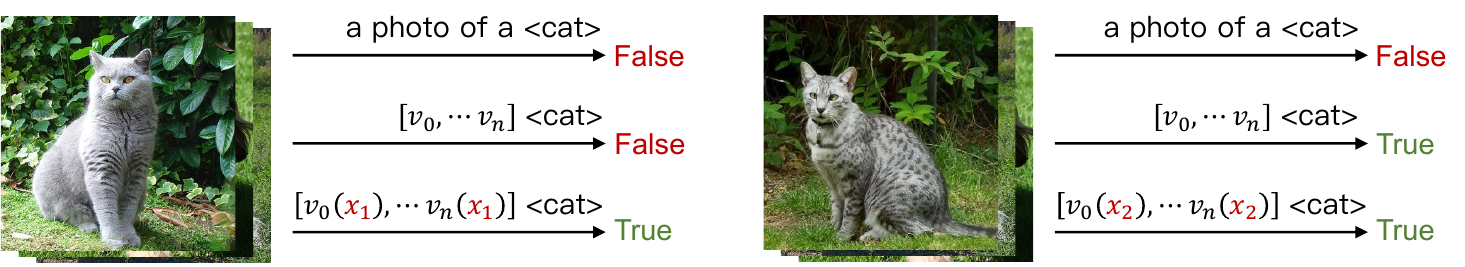}
    \caption{Inference with different prompts. \textbf{Top}: inference with hand-crafted prompts fails to recognize the class `cat'; \textbf{Middle}: Inference with fixed prompts tuned by APT methods cannot recognize all adversarial images; \textbf{Bottom}: Inference with test-time prompts optimized for each image produces accurate recognitions.}
    \label{fig1}
\end{figure}

Despite its promising results, APT faces three fundamental limitations: (1) The distribution-dependent nature of learnable prompts restricts their generalization to out-of-distribution scenarios and novel tasks. (2) The requirement for task-specific annotated data presents significant challenges for zero-shot applications. (3) While APT methods improve robustness on specific tasks~\cite{stutz2019disentangling}, they often compromise the model's overall generalization performance~\cite{su2018robustness, pedraza2021relationship}. 
Given that real-world applications involve an unbounded set of potential tasks, evaluating adversarial robustness solely on predefined downstream tasks proves insufficient. Therefore, achieving zero-shot adversarial robustness—maintaining performance against adversarial attacks on unseen tasks without task-specific training—remains an open challenge.

The key to addressing the zero-shot adversarial robustness challenge lies in adaptively identifying a robust prompt that consistently aligns an adversarial image embedding with its correct text embedding for each test sample. To this end, we propose a simple yet effective framework called \textbf{Test-Time Adversarial Prompt Tuning (TAPT)}, which dynamically tunes a robust prompt on the fly based on only the provided test sample.
Specifically, TAPT learns defensive prompts by minimizing two unsupervised losses: (1) multi-view entropy, which ensures consistent predictions across various augmented views of each test sample, and (2) adversarial-clean embedding alignment, which aligns the means and variances of test sample embeddings with pre-computed adversarial-clean embeddings of a public dataset (e.g., ImageNet) to enhance inference robustness. Notably, TAPT operates during the inference phase without requiring any task-specific training set or annotations.  The different inference schemes of training-time defense APT methods and our test-time defense TAPT are illustrated in Figure~\ref{fig1}.

We evaluated TAPT against both white-box and black-box adversarial attacks across 11 benchmark datasets. TAPT demonstrated superior performance compared to vanilla CLIP (using hand-crafted prompts) and existing APT methods across three different prompt designs: visual-only prompts, visual-language (V-L) joint prompts, and V-L independent prompts. By dynamically adapting prompts at inference time, TAPT opens a new direction for safeguarding the inference process of pre-trained VLMs. Offering a flexible and efficient solution for zero-shot adversarial robustness while maintaining performance on clean samples, our TAPT method bridges the gap between the need for adversarial robustness and the performance challenges posed by open, real-world environments.

In summary, our main contributions are:
\begin{itemize}
    \item We propose a novel test-time defense method named \textbf{Test-Time Adversarial Prompt Tuning (TAPT)} to robustify the zero-shot inference process of pre-trained VLMs. 
    TAPT adapts the prompt for each test image to achieve robust inference without the need for task-specific tuning for downstream datasets.
    To the best of our knowledge, TAPT is the first inference-time adversarial defense method for pre-trained VLMs.

    \item TAPT introduces an adversarial-clean alignment loss that aligns the distribution of a test sample with pre-computed adversarial-clean distributions from a public dataset (ImageNet), thereby improving adversarial robustness while maintaining accuracy on clean samples. It also leverages the advantage of APT by using pre-tuned prompts on ImageNet to further enhance zero-shot adversarial robustness.
   
    \item We conduct extensive experiments on 11 datasets, including ImageNet and 10 other zero-shot datasets. Our results demonstrate that TAPT significantly outperforms existing APT baselines against both white-box and black-box attacks. Specifically, TAPT improves zero-shot adversarial robustness against AutoAttack by 36.6\% with ViT-B/16, and by 38.0\% with ViT-B/32, respectively.
\end{itemize}
\section{Related Work}
\label{sec:related}

Here, we briefly review related works on adversarial attacks and defenses for pre-trained VLMs, and test-time adaptation techniques proposed to improve generalization.

\vspace{0.1cm}
\noindent\textbf{Adversarial Attacks on Pre-trained VLMs}  
Adversarial attacks on pre-trained VLMs are broadly categorized as white-box or black-box attacks according to the threat model. In white-box attacks, the attacker has full access to the model parameters and can directly compute adversarial gradients~\cite{madry2017towards, zhang2022towards, zhou2023advclip, ma2024imbalanced}. Black-box attacks, on the other hand, restrict the attacker to querying model outputs~\cite{xie2019improving, zhao2024evaluating, lu2023set, he2023sa, wang2024transferable, yin2024vlattack, fang2024one, zhang2024universal}. Traditional single-modal attacks designed for vision models, such as PGD~\cite{madry2017towards}, DI~\cite{xie2019improving}, and AutoAttack~\cite{croce2020reliable}, can be used directly to attack the image encoders of pre-trained VLMs. Several recent multi-modal attacks have targeted pre-trained VLMs by simultaneously exploiting vulnerabilities in both their image and text encoders. For example, Co-Attack~\cite{zhang2022towards} pioneered white-box multi-modal attacks, perturbing both image and text modalities concurrently. SGA~\cite{lu2023set} extended Co-Attack to the black-box setting, improving the transferability of multi-modal adversarial examples. VLATTACK~\cite{yin2024vlattack} further generates adversarial examples by fusing perturbations of images and texts from both single-modal and multi-modal.

\begin{figure*}[htbp]
    \centering
    \includegraphics[width=1\linewidth]{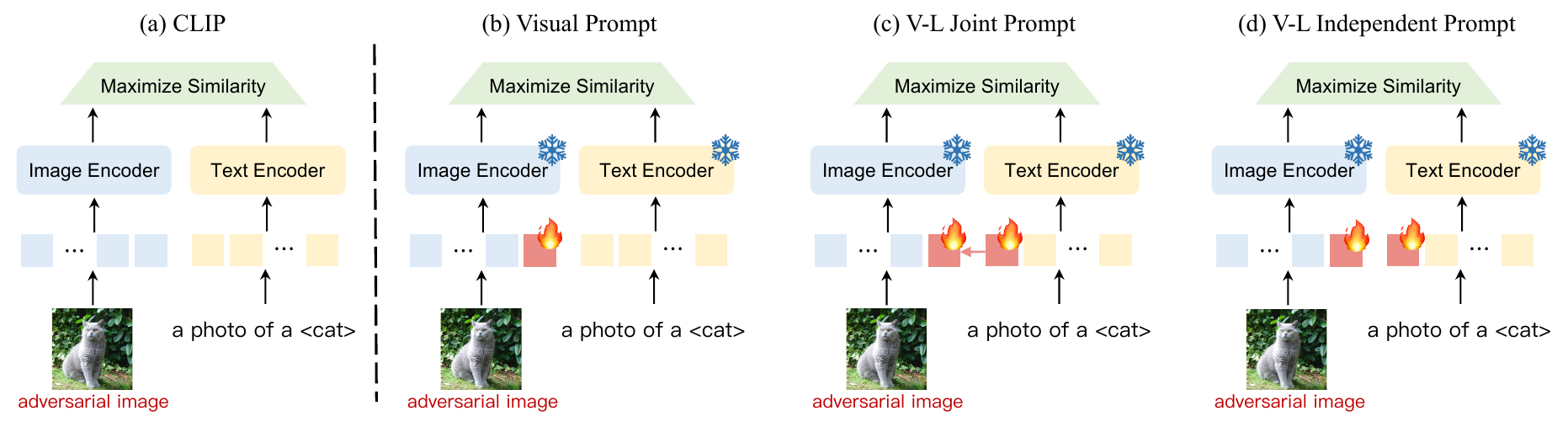}
    \caption{An illustration of CLIP and different adversarial prompt tuning schemes. (a) The original CLIP~\cite{radford2021learning}; (b) - (d) Adversarial prompt tuning with three distinctive prompt designs: Visual Prompt (b), V-L Joint Prompt (c), and V-L Independent Prompt (d).}
    \label{fig2}
\end{figure*}

\vspace{0.1cm}
\noindent\textbf{Adversarial Defenses for Pre-trained VLMs} 
Adversarial training/tuning is a widely used defense strategy for pre-trained VLMs, with existing methods generally classified into adversarial contrastive tuning~\cite{wang2024revisiting, mao2023understanding, schlarmannrobust, wang2024pre, zhou2024revisiting, wang2024advqdet} and adversarial prompt tuning~\cite{zhang2023adversarial, li2024one, zhou2024few}. Adversarial contrastive tuning focuses on enhancing the adversarial robustness of the backbone model. For example, TeCoA~\cite{mao2023understanding} examines the impact of fine-tuning and visual prompt tuning on the zero-shot adversarial robustness of VLMs. FARE~\cite{schlarmannrobust} improves CLIP's image encoder through unsupervised adversarial fine-tuning, enhancing adversarial robustness in models like LLaVA and OpenFlamingo without retraining. PMG-AFT~\cite{wang2024pre} introduces an auxiliary branch to boost zero-shot adversarial robustness, while MMCoA~\cite{zhou2024revisiting} investigates VLM vulnerabilities to multimodal attacks. In contrast, AdvPT~\cite{zhang2023adversarial} and APT~\cite{li2024one} offer an efficient approach to bolster the adversarial robustness of VLMs by tuning only the textual prompts without altering the model parameters. FAP~\cite{zhou2024few} further refines APT by balancing cross-modal consistency between benign and adversarial inputs. While these methods are all training-time defense methods that need to pre-tune the prompt for a specific downstream task, in this work we propose a novel test-time defense method that optimizes the prompt on the fly during inference and is task-agnostic.

\vspace{0.1cm}
\noindent\textbf{Test-Time Adaptation}  
Test-time adaptation (TTA) methods enhance the generalization performance of pre-trained models by adapting to individual test samples or batches, addressing distribution shifts between training (source) and testing (target) data. Early TTA methods~\cite{schneider2020improving, nado2020evaluating} address domain shift by updating batch normalization statistics based on test batch statistics. Building on this, TENT~\cite{wangtent} improves adaptation by updating batch normalization layers to minimize the entropy of predicted probabilities for each test batch, while MEMO~\cite{zhang2022memo} extends this by minimizing entropy over multiple augmented input samples. Inspired by TENT, subsequent TTA methods such as CoTTA~\cite{wang2022continual} and EATA~\cite{niu2022efficient} enable pre-trained models to adapt to continuously shifting test distributions. More recently, methods like TPT~\cite{shu2022test} and PromptAlign~\cite{abdul2024align} have focused on tuning prompts exclusively at test time to ensure consistent predictions across various augmented views of a test sample. 
Kim \etal~\cite{kimtest} finds that combining TTA with AGL~\cite{baek2022agreement} further improves out-of-distribution (OOD) accuracy prediction. In contrast to existing TTA methods which primarily emphasize performance on clean samples, in this work we introduce TAPT, a test-time defense, that specifically enhances zero-shot adversarial robustness against potential attacks.

\section{Proposed Method}
\label{sec:method}

\subsection{Preliminaries}
\noindent\textbf{Threat Model}
We assume a white-box threat model in which the adversary has full knowledge of the target model's architecture and parameters, allowing direct perturbation of test images based on adversarial gradients prior to inference. The defender, or model owner, can deploy any defense strategies to protect against potential adversarial attacks. Specifically, we focus on securing CLIP zero-shot inference, where the defender lacks access to task-specific training data or annotations of the downstream application.

\vspace{0.1cm}
\noindent\textbf{CLIP}
We denote the CLIP image encoder as $\mathcal{I}$, parameterized by $\theta_{\mathcal{I}}$, and the text encoder as $\mathcal{T}$, parameterized by $\theta_{\mathcal{T}}$. Considering a $K$-class classification problem, where each image $\rvx$ is associated with a class label in the format $\texttt{"a photo of a <class>"}$. For a clean sample $\rvx \in [0, 1]^d$ and a target CLIP model comprising $\{\mathcal{I},\mathcal{T}\}$, a white-box adversarial attack seeks to generate an adversarial example $\rvx'$ that maximizes the model loss as follows:
\begin{equation}
    \rvx' = \argmax_{\| \rvx' - \rvx \|_{\infty} \leq \epsilon} \Ls(\mathcal{I}(\rvx'), \mathcal{T}(y)),
\end{equation}
where $\Ls(\cdot)$ represents the loss function, $\rvx'$ is the adversarial example, and $\epsilon$ denotes the perturbation budget.

\begin{figure*}[htbp]
    \centering
    \includegraphics[width=1\linewidth]{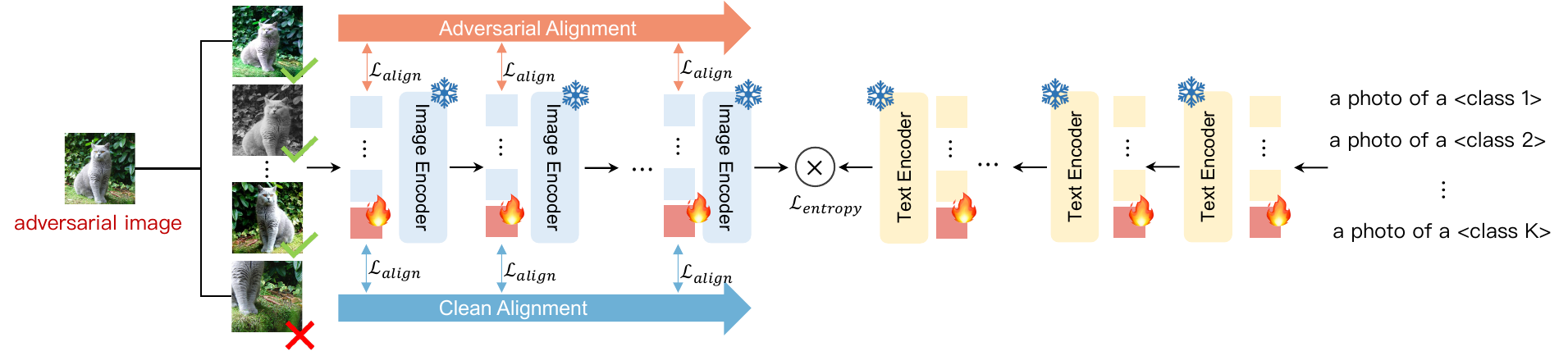}
    \caption{An overview of our proposed TAPT method: Given an adversarial image, TAPT generates multiple augmented views of the image and retains only those views with low entropy in their averaged prediction probabilities. During inference, TAPT then optimizes the prompt by minimizing multi-view entropy across these selected views while aligning their embedding distribution with pre-computed adversarial-clean statistics from a public dataset (ImageNet).}
    \label{fig3}
\end{figure*}

\vspace{0.1cm}
\noindent\textbf{Adversarial Prompt Tuning (APT)}
APT applies adversarial training during the prompt tuning process to enhance adversarial robustness, with recent APT methods primarily developed to defend the visual component of CLIP. Rather than relying on hand-crafted prompts like $\texttt{"a photo of a <class>"}$, APT learns robust prompts from training data to improve adversarial robustness on downstream tasks. As shown in Figure~\ref{fig2}, APT can be extended to three distinctive prompt designs: (1) Visual-only (prompts applied solely to the vision branch), (2) Vision-Language (V-L) joint (shared prompts across both branches), and (3) V-L independent (separate prompts for each branch). Specifically, APT optimizes the prompts $\bm{P} = \{\bm{P_v}, \bm{P_t}\} \in \mathbb{R}^{L \times D}$, where $\bm{P_v}$ and $\bm{P_t}$ are the visual and textual prompts, respectively, $L$ denotes the number of prompt tokens, and $D$ is the embedding dimension.

Given a downstream training dataset $\mathcal{D}_{\text{train}} = \{(\rvx,y)\}$, the learnable visual prompt $\bm{P_v}$ is appended to the visual input tokens, forming the sequence $\{\rvx, \bm{P_v}\} = \{\textsc{cls}, e_1, e_2, \cdots, e_M, \bm{P_v}\}$. Similarly, the textual prompt $\bm{P_t}$ is appended to the text input, forming $\{y, \bm{P_t}\}$. APT enhances adversarial robustness by aligning the embeddings of clean text with those of adversarial images through a min-max optimization. The corresponding optimization problem can be formulated as: 
\begin{equation}
\label{eq:apt}
   \argmin_{\bm{P}} \mathbb{E}_{\mathcal{D}_{\text{train}}} \max_{\|\rvx' - \rvx\|_{\infty} \leq \epsilon} \Ls(\mathcal{I}(\rvx^{'}, \bm{P_v}), \mathcal{T}(y, \bm{P_t})),
\end{equation}
where $\mathcal{I}(\rvx^{'}, \bm{P_v})$ and $\mathcal{T}(y, \bm{P_t})$ denote the adversarial image embedding and text embedding, respectively. APT methods tune the prompts on the training dataset of each downstream task. At test time, the tuned prompts are fixed to perform inference for different test images.

\begin{algorithm}[t]
    \caption{Test-Time Adversarial Prompt Tuning}
    \label{algorithm}
    \begin{algorithmic}[1]
    \State {\bfseries Input:} input image $\rvx$, image encoder $\mathcal{I}$, text encoder $\mathcal{T}$, augmentation function $\mathcal{A}$, entropy threshold $\tau$, pre-computed $\mathcal{D}_{\text{public}}$ statistics $\{\mu_{\text{adv}}, \sigma_{\text{adv}}, \mu_{\text{clean}}, \sigma_{\text{clean}}\}$
    \State {\bfseries Output:} Learnable prompts $\bm{P}$
    
    \State \textbf{1. Initialize adversarial prompts}:
    \State \quad $\bm{P} \gets \text{APT}(\mathcal{D}_{\text{public}}, \epsilon; \mathcal{I}, \mathcal{T})$
    
    \State \textbf{2. Multi-view entropy based sample selection:}
    \State \quad Select the top $\tau$ entropy from $\mathcal{A}(\mathbf{x})$ to form $\mathcal{H}_\tau(\mathbf{x})$.
    
    \State \textbf{4. Compute multi-view entropy loss:}
    \State \quad Calculate $\mathcal{L}_{\text{entropy}}$ over selected views $\mathcal{H}_\tau(\rvx)$
    
    \State \textbf{5. Compute current embedding statistics:}
    \State \quad \textbf{for} each layer $l$ in $\mathcal{I}$ \textbf{do}
        \State \qquad $\mu_l = \text{mean}(I_l(\hat{\rvx}, \bm{P}))$ for $\hat{\rvx}$ in $\mathcal{H}_\tau(\rvx)$
        \State \qquad $\sigma_l^2 = \text{std}(\mathcal{I}_l(\hat{\rvx}, \bm{P}))$ for $\hat{\rvx}$ in $\mathcal{H}_\tau(\rvx)$
    \State \quad \textbf{end for}
    
    \State \textbf{6. Adversarial-clean embedding alignment:}
    \State \quad \scalebox{0.95}{$\mathcal{L}_{\text{adv}} = \dfrac{1}{L} \sum\limits_{l=1}^L \left( \left\| \mu_l - \mu_{\text{adv}, l} \right\|_1 + \left\| \sigma^2_l - \sigma^2_{\text{adv}, l} \right\|_1 \right)$}
    \State \quad \scalebox{0.95}{$\mathcal{L}_{\text{clean}} = \dfrac{1}{L} \sum\limits_{l=1}^L \left( \left\| \mu_l - \mu_{\text{clean}, l} \right\|_1 + \left\| \sigma^2_l - \sigma^2_{\text{clean}, l} \right\|_1 \right)$}
    
    \State \textbf{7. Optimize prompts:}
    \State \quad $\mathcal{L}_{\text{TAPT}} = \mathcal{L}_{\text{entropy}} + \alpha \mathcal{L}_{\text{adv}} + (1-\alpha) \mathcal{L}_{\text{clean}}$
    \State \quad Optimize $\bm{P} \leftarrow$ Minimize $\mathcal{L}_{\text{TAPT}}$
    
    \end{algorithmic}
\end{algorithm}

\subsection{Test-Time Adversarial Prompt Tuning (TAPT)}
\noindent\textbf{Framework Overview}
As illustrated in Figure~\ref{fig3}, TAPT comprises two main modules: (1) multi-view entropy-based sample selection, and (2) adversarial-clean embedding alignment. The defense procedure of TAPT operates as follows. Given a test image $\rvx \in \mathcal{D}_{\text{test}}$, TAPT begins by generating $M$ randomly augmented views $\{\mathcal{A}_1(\rvx), \mathcal{A}_2(\rvx), \dots, \mathcal{A}_M(\rvx)\}$ through random augmentations $\mathcal{A}$. The multi-view entropy-based sample selection module then chooses the top-$K$ views with the lowest entropy in their averaged prediction probabilities. Using these selected views, TAPT optimizes the prompt $\bm{P}$ during inference by minimizing multi-view entropy and enforcing adversarial-clean alignment. The prompt is reset to its initial state before processing each new test sample or batch. The complete procedure of TAPT is outlined in \cref{algorithm}.

\vspace{0.1cm}
\noindent\textbf{Multi-View Entropy-Based Sample Selection}
Following prior works~\cite{shu2022test, abdul2024align}, we first discard ineffective augmented views $\mathcal{A}(\rvx)$ (e.g., when a random crop removes essential image content) by applying a selection filter with a threshold $\tau$. This filter retains only augmented views with low entropy (high-confidence predictions). Specifically, we define the set of selected augmented views as $\mathcal{H}_\tau(\rvx) = \{\mathcal{A}_j(\rvx) | 1 \le j \le M, H(\mathcal{A}_j(\rvx)) \le H_{\tau}\}$, where $H_{\tau}$ is the entropy threshold corresponding to the top $\tau$ lowest entropy values among all $M$ augmentations. TAPT then optimizes prompts by minimizing the multi-view entropy of the averaged prediction probability over these selected views:
\begin{equation}
\label{eq:entropy}
    \Ls_{\text{entropy}} = -\sum_{i=1}^{K} \Tilde{p}(y_i|\mathcal{H}_\tau(\rvx),\bm{P}) \log \Tilde{p}(y_i|\mathcal{H}_\tau(\rvx),\bm{P}),
\end{equation}
where $\tilde{p}(y_i|\mathcal{H}_\tau(\rvx), \mathbf{P}) = \frac{1}{|\mathcal{H}_\tau(\rvx)|} \sum_{\hat{\rvx} \in \mathcal{H}_\tau(\rvx)} p(y_i|\hat{\rvx}, \mathbf{P})$ denotes the average predicted probability for class $y_i$ over the selected augmented views $\mathcal{H}_\tau(\rvx)$ when using prompt $\bm{P}$. This optimization encourages the model to achieve consistent predictions by adjusting the prompt specifically for the given instance.

\vspace{0.1cm}
\noindent\textbf{Adversarial-Clean Embedding Alignment} 
An adversarial test image $\rvx$ can shift the image embeddings produced by the image encoder $\mathcal{I}(\rvx, \bm{P})$ compared to those from clean images, potentially misleading the model. To mitigate this vulnerability, we align the mean and variance of the test image embedding with pre-computed statistics from a public dataset $\mathcal{D}_{\text{public}}$. Ideally, this public dataset would originate from the original CLIP pre-training data. However, since this data is unavailable, we use ImageNet as a proxy, given CLIP's strong zero-shot performance on ImageNet with standard prompt tuning~\cite{bahng2022exploring}. Specifically, we compute the mean and variance of the current embeddings for alignment as follows:
\begin{align}
\label{eq:dist}
    \mu_l(\mathcal{H}_\tau(\rvx);\bm{P}) &= \frac{1}{|\mathcal{H}_\tau(\rvx)|} \sum_{\hat{\rvx} \in \mathcal{H}_\tau(\rvx)} \mathcal{I}_l(\hat{\rvx},\bm{P}), \\
    \sigma^2_l(\mathcal{H}_\tau(\rvx);\bm{P}) &= \frac{\sum_{\hat{\rvx} \in \mathcal{H}_\tau(\rvx)} (\mathcal{I}_l(\hat{\rvx},\bm{P}) - \mu_l(\mathcal{H}_\tau(\rvx);\bm{P}))^2}{|\mathcal{H}_\tau(\rvx)| - 1}
\end{align}
where $\mathcal{I}_l(\hat{\rvx},\bm{P})$ represents the embedding vector at layer $l$ for the augmented input $\hat{\rvx} \in \mathcal{H}_\tau(\rvx)$ given prompt $\bm{P}$. Here, $\mu_l(\mathcal{H}_\tau(\rvx);\bm{P})$ and $\sigma^2_l(\mathcal{H}_\tau(\rvx);\bm{P})$ denote the mean and variance of the test sample embeddings at layer $l$, respectively. We similarly pre-compute the mean and variance of embeddings from the public dataset using the robust prompt $\bm{P_{adv}}$ (obtained via APT on the public data) and clean prompt $\bm{P_{clean}}$ (obtained via standard prompt tuning on the public data). These offline statistics are represented by $\mu_{\text{adv}}$, $\sigma^2_{\text{adv}}$, $\mu_{\text{clean}}$, and $\sigma^2_{\text{clean}}$, respectively. We then align the mean and variance of the current embeddings with these pre-computed statistics as follows:
\begin{align}
\label{eq:align}
    \Ls_\text{adv} &= \frac{1}{L} \sum_{l=1}^{L} (\lVert \mu_l - \mu_{adv,l}\rVert_1 + \lVert \sigma^2_l - \sigma^2_{adv,l} \rVert_1),   \\
    \Ls_\text{clean} &= \frac{1}{L} \sum_{l=1}^{L} (\lVert \mu_l - \mu_{clean,l} \rVert_1 + \lVert \sigma^2_l - \sigma^2_{clean,l} \rVert_1),   \\
    \Ls_\text{TAPT} &= \Ls_{\text{entropy}} + \alpha \Ls_\text{adv} + (1-\alpha) \Ls_\text{clean}, 
\end{align}
where $\alpha$ is a hyperparameter. Setting $\alpha=0$ aligns the prompt with the clean distribution, which may reduce adversarial robustness, while setting $\alpha=1$ aligns the prompt with the robust distribution, which may impact performance on clean samples. Our final objective combines the multi-view entropy loss with the adversarial-clean alignment to optimize the prompt during inference for a given test sample. This approach enhances adversarial robustness while preserving accuracy on clean samples. \textit{\textbf{Note that, to ensure inference efficiency, TAPT performs only a single step of prompt tuning for each inference.}}

\section{Experiments}
\label{sec:exp}

\begin{table*}[htp]
\centering
\resizebox{1.0\linewidth}{!}{
\setlength{\tabcolsep}{1.0mm}{
\begin{tabular}{llllllllllllllllll}
\toprule
\multicolumn{4}{l}{} & \textbf{ImageNet} & \textbf{Caltech101} & \textbf{DTD} & \textbf{EuroSAT} & \textbf{Pets} & \textbf{Aircraft} & \textbf{Food101} & \textbf{Flowers} & \textbf{Cars} & \textbf{SUN397} & \textbf{UCF101} & \textbf{Avg.} \\
\midrule
\multirow{6}{*}{\rotatebox{90}{\textbf{CLIP}}} & 
\multirow{6}{*}{Vanilla} &
\multirow{3}{*}{ViT-B/16} &
PGD & 1.4 & 21.3 & 1.4 & 6.0 & 5.0 & 0.0 & 9.8 & 1.6 & 0.7 & 0.8 & 1.8 & 4.5 \\
&&& DI & 6.1 & 25.8 & 8.7 & 0.3 & 11.1 & 0.5 & 9.5 & 3.0 & 3.4 & 4.9 & 4.7 & 7.1 \\
&&& AA & 0.0 & 0.0 & 0.0 & 0.1 & 0.0 & 0.1 & 0.0 & 0.0 & 0.0 & 0.0 & 0.0 & 0.1 \\
\cline{3-16}
&&\multirow{3}{*}{ViT-B/32} &
PGD & 1.3 & 22.9 & 5.0 & 0.0 & 2.6 & 0.0 & 3.6 & 1.5 & 0.2 & 1.1 & 1.7 & 3.6 \\
&&& DI & 6.4 & 37.2 & 11.1 & 0.7 & 9.9 & 0.1 & 7.5 & 9.3 & 3.9 & 8.6 & 5.8 & 9.1 \\
&&& AA & 0.0 & 0.0 & 0.0 & 0.1 & 0.0 & 0.1 & 0.0 & 0.1 & 0.1 & 0.0 & 0.1 & 0.1 \\
\midrule
\multirow{12}{*}{\rotatebox{90}{\textbf{Visual Only}}} & 
\multirow{6}{*}{APT-V} &
\multirow{3}{*}{ViT-B/16} &
PGD & 19.4 & 61.2 & 18.5 & 8.0 & 37.4 & 3.9 & 12.1 & 25.1 & 9.8 & 17.2 & 16.4 & 20.8 \\
&&& DI & 29.2 & 67.5 & 23.5 & 12.3 & 47.1 & 5.0 & 21.8 & 30.2 & 18.1 & 27.1 & 22.0 & 27.6 \\
&&& AA & 14.8 & 55.9 & 15.2 & 2.3 & 31.3 & 2.3 & 8.3 & 18.0 & 5.8 & 12.4 & 12.7 & 16.3 \\
\cline{3-16}
&&\multirow{3}{*}{ViT-B/32} &
PGD & 18.9 & 63.8 & 20.2 & 0.6 & 36.1 & 2.7 & 15.0 & 23.1 & 9.4 & 19.1 & 18.4 & 20.7 \\
&&& DI & 26.9 & 68.6 & 22.0 & 5.1 & 46.0 & 4.8 & 23.7 & 26.5 & 13.9 & 26.4 & 24.6 & 26.2 \\
&&& AA & 8.3 & 44.9 & 12.7 & 0.1 & 16.4 & 0.5 & 5.5 & 9.0 & 2.7 & 7.5 & 7.8 & 10.5 \\
\cline{2-16}
& 
\multirow{6}{*}{\textbf{TAPT-V}} &
\multirow{3}{*}{ViT-B/16} &
PGD & 40.1 {\small \color{deepgreen}(20.7$\uparrow$)} & 69.7 {\small \color{deepgreen}(8.5$\uparrow$)} & 28.1 {\small \color{deepgreen}(9.6$\uparrow$)} & 23.9 {\small \color{deepgreen}(15.9$\uparrow$)} & 49.2 {\small \color{deepgreen}(11.8$\uparrow$)} & 11.5 {\small \color{deepgreen}(7.6$\uparrow$)} & 54.5 {\small \color{deepgreen}(42.4$\uparrow$)} & 41.1 {\small \color{deepgreen}(16.0$\uparrow$)} & 28.6 {\small \color{deepgreen}(18.8$\uparrow$)} & 40.3 {\small \color{deepgreen}(23.1$\uparrow$)} & 36.5 {\small \color{deepgreen}(20.1$\uparrow$)} & 38.5 {\small \color{deepgreen}(17.7$\uparrow$)} \\
&&& DI & 46.7 {\small \color{deepgreen}(17.5$\uparrow$)} & 75.8 {\small \color{deepgreen}(8.3$\uparrow$)} & 33.2 {\small \color{deepgreen}(9.7$\uparrow$)} & 33.0 {\small \color{deepgreen}(20.7$\uparrow$)} & 56.7 {\small \color{deepgreen}(9.6$\uparrow$)} & 12.7 {\small \color{deepgreen}(7.7$\uparrow$)} & 61.5 {\small \color{deepgreen}(39.7$\uparrow$)} & 46.2 {\small \color{deepgreen}(16.0$\uparrow$)} & 35.7 {\small \color{deepgreen}(17.6$\uparrow$)} & 45.5 {\small \color{deepgreen}(18.4$\uparrow$)} & 40.1 {\small \color{deepgreen}(18.1$\uparrow$)} & 44.3 {\small \color{deepgreen}(16.7$\uparrow$)} \\
&&& AA & 49.2 {\small \color{deepgreen}(34.4$\uparrow$)} & 75.7 {\small \color{deepgreen}(19.8$\uparrow$)} & 36.6 {\small \color{deepgreen}(21.4$\uparrow$)} & 36.9 {\small \color{deepgreen}(34.6$\uparrow$)} & 57.1 {\small \color{deepgreen}(25.8$\uparrow$)} & 19.4 {\small \color{deepgreen}(17.1$\uparrow$)} & 68.8 {\small \color{deepgreen}(60.5$\uparrow$)} & 52.9 {\small \color{deepgreen}(34.9$\uparrow$)} & 44.0 {\small \color{deepgreen}(38.2$\uparrow$)} & 48.9 {\small \color{deepgreen}(36.5$\uparrow$)} & 48.1 {\small \color{deepgreen}(35.4$\uparrow$)} & 48.9 {\small \color{deepgreen}(32.6$\uparrow$)} \\
\cline{3-16}
&&\multirow{3}{*}{ViT-B/32} &
PGD & 42.2 {\small \color{deepgreen}(23.3$\uparrow$)} & 79.4 {\small \color{deepgreen}(15.6$\uparrow$)} & 32.2 {\small \color{deepgreen}(12.0$\uparrow$)} & 24.4 {\small \color{deepgreen}(23.8$\uparrow$)} & 62.4 {\small \color{deepgreen}(26.3$\uparrow$)} & 12.1 {\small \color{deepgreen}(9.4$\uparrow$)} & 53.2 {\small \color{deepgreen}(38.2$\uparrow$)} & 47.1 {\small \color{deepgreen}(24.0$\uparrow$)} & 33.5 {\small \color{deepgreen}(24.1$\uparrow$)} & 44.1 {\small \color{deepgreen}(25.0$\uparrow$)} & 44.5 {\small \color{deepgreen}(26.1$\uparrow$)} & 43.2 {\small \color{deepgreen}(22.5$\uparrow$)} \\
&&& DI & 46.5 {\small \color{deepgreen}(19.6$\uparrow$)} & 81.5 {\small \color{deepgreen}(12.9$\uparrow$)} & 33.6 {\small \color{deepgreen}(11.6$\uparrow$)} & 24.8 {\small \color{deepgreen}(19.7$\uparrow$)} & 66.0 {\small \color{deepgreen}(20.0$\uparrow$)} & 13.3 {\small \color{deepgreen}(8.5$\uparrow$)} & 57.7 {\small \color{deepgreen}(34.0$\uparrow$)} & 48.7 {\small \color{deepgreen}(22.2$\uparrow$)} & 37.5 {\small \color{deepgreen}(23.6$\uparrow$)} & 47.3 {\small \color{deepgreen}(20.9$\uparrow$)} & 47.7 {\small \color{deepgreen}(23.1$\uparrow$)} & 45.9 {\small \color{deepgreen}(19.7$\uparrow$)} \\
&&& AA & 44.1 {\small \color{deepgreen}(35.8$\uparrow$)} & 76.3 {\small \color{deepgreen}(31.4$\uparrow$)} & 33.5 {\small \color{deepgreen}(20.8$\uparrow$)} & 31.4 {\small \color{deepgreen}(31.3$\uparrow$)} & 54.9 {\small \color{deepgreen}(38.5$\uparrow$)} & 16.4 {\small \color{deepgreen}(15.9$\uparrow$)} & 53.6 {\small \color{deepgreen}(48.1$\uparrow$)} & 47.3 {\small \color{deepgreen}(38.3$\uparrow$)} & 42.4 {\small \color{deepgreen}(39.7$\uparrow$)} & 45.6 {\small \color{deepgreen}(38.1$\uparrow$)} & 45.0 {\small \color{deepgreen}(37.2$\uparrow$)} & 44.6 {\small \color{deepgreen}(34.1$\uparrow$)} \\
\midrule
\multirow{12}{*}{\rotatebox{90}{\textbf{V-L Joint}}} & 
\multirow{6}{*}{APT-VLJ} &
\multirow{3}{*}{ViT-B/16} &
PGD & 23.9 & 61.7 & 18.7 & 9.7 & 41.4 & 3.2 & 14.1 & 23.4 & 12.2 & 17.7 & 15.5 & 22.0 \\
&&& DI & 34.9 & 72.0 & 25.0 & 10.5 & 52.8 & 4.2 & 24.6 & 32.9 & 18.9 & 28.5 & 24.8 & 29.9 \\
&&& AA & 16.5 & 53.2 & 12.6 & 5.6 & 31.3 & 1.7 & 8.0 & 16.0 & 4.9 & 11.1 & 11.1 & 15.6 \\
\cline{3-16}
&&\multirow{3}{*}{ViT-B/32} &
PGD & 21.4 & 64.3 & 14.7 & 10.4 & 37.7 & 2.0 & 16.5 & 21.0 & 8.9 & 17.5 & 18.0 & 21.1 \\
&&& DI & 30.4 & 69.5 & 17.0 & 11.9 & 47.9 & 2.9 & 26.7 & 26.1 & 17.3 & 26.5 & 26.4 & 27.5 \\
&&& AA & 10.3 & 46.2 & 10.0 & 3.0 & 18.5 & 0.6 & 6.7 & 9.5 & 1.7 & 7.4 & 7.2 & 11.1 \\
\cline{2-16}
& 
\multirow{6}{*}{\textbf{TAPT-VLJ}} &
\multirow{3}{*}{ViT-B/16} &
PGD & 50.2 {\small \color{deepgreen}(26.3$\uparrow$)} & 81.0 {\small \color{deepgreen}(19.3$\uparrow$)} & 29.5 {\small \color{deepgreen}(10.8$\uparrow$)} & 13.5 {\small \color{deepgreen}(3.8$\uparrow$)} & 68.7 {\small \color{deepgreen}(27.3$\uparrow$)} & 5.6 {\small \color{deepgreen}(2.4$\uparrow$)} & 41.7 {\small \color{deepgreen}(27.6$\uparrow$)} & 39.3 {\small \color{deepgreen}(15.9$\uparrow$)} & 28.1 {\small \color{deepgreen}(15.9$\uparrow$)} & 39.8 {\small \color{deepgreen}(22.1$\uparrow$)} & 41.5 {\small \color{deepgreen}(26.0$\uparrow$)} & 39.9 {\small \color{deepgreen}(17.9$\uparrow$)} \\
&&& DI & 51.7 {\small \color{deepgreen}(16.8$\uparrow$)} & 82.3 {\small \color{deepgreen}(10.3$\uparrow$)} & 30.4 {\small \color{deepgreen}(5.4$\uparrow$)} & 14.5 {\small \color{deepgreen}(4.0$\uparrow$)} & 69.5 {\small \color{deepgreen}(16.7$\uparrow$)} & 5.8 {\small \color{deepgreen}(1.6$\uparrow$)} & 43.2 {\small \color{deepgreen}(18.6$\uparrow$)} & 40.2 {\small \color{deepgreen}(7.3$\uparrow$)} & 30.2 {\small \color{deepgreen}(11.3$\uparrow$)} & 41.3 {\small \color{deepgreen}(12.8$\uparrow$)} & 42.2 {\small \color{deepgreen}(17.4$\uparrow$)} & 41.0 {\small \color{deepgreen}(11.1$\uparrow$)} \\
&&& AA & 52.4 {\small \color{deepgreen}(35.9$\uparrow$)} & 81.5 {\small \color{deepgreen}(28.3$\uparrow$)} & 30.2 {\small \color{deepgreen}(17.6$\uparrow$)} & 15.4 {\small \color{deepgreen}(9.8$\uparrow$)} & 69.1 {\small \color{deepgreen}(37.8$\uparrow$)} & 5.8 {\small \color{deepgreen}(4.1$\uparrow$)} & 44.8 {\small \color{deepgreen}(36.8$\uparrow$)} & 40.0 {\small \color{deepgreen}(24.0$\uparrow$)} & 31.0 {\small \color{deepgreen}(26.1$\uparrow$)} & 42.1 {\small \color{deepgreen}(31.0$\uparrow$)} & 44.3 {\small \color{deepgreen}(33.2$\uparrow$)} & 41.5 {\small \color{deepgreen}(25.9$\uparrow$)} \\
\cline{3-16}
&&\multirow{3}{*}{ViT-B/32} &
PGD & 52.1 {\small \color{deepgreen}(30.7$\uparrow$)} & 80.6 {\small \color{deepgreen}(16.3$\uparrow$)} & 27.1 {\small \color{deepgreen}(12.4$\uparrow$)} & 13.4 {\small \color{deepgreen}(3.0$\uparrow$)} & 72.0 {\small \color{deepgreen}(34.3$\uparrow$)} & 7.6 {\small \color{deepgreen}(5.6$\uparrow$)} & 52.5 {\small \color{deepgreen}(36.0$\uparrow$)} & 39.9 {\small \color{deepgreen}(18.0$\uparrow$)} & 32.3 {\small \color{deepgreen}(23.4$\uparrow$)} & 44.6 {\small \color{deepgreen}(27.1$\uparrow$)} & 45.3 {\small \color{deepgreen}(27.1$\uparrow$)} & 42.4 {\small \color{deepgreen}(21.3$\uparrow$)} \\
&&& DI & 52.4 {\small \color{deepgreen}(22.0$\uparrow$)} & 80.5 {\small \color{deepgreen}(11.0$\uparrow$)} & 27.1 {\small \color{deepgreen}(10.1$\uparrow$)} & 13.6 {\small \color{deepgreen}(1.7$\uparrow$)} & 71.6 {\small \color{deepgreen}(23.7$\uparrow$)} & 7.7 {\small \color{deepgreen}(4.8$\uparrow$)} & 52.0 {\small \color{deepgreen}(25.3$\uparrow$)} & 40.0 {\small \color{deepgreen}(13.9$\uparrow$)} & 33.4 {\small \color{deepgreen}(16.1$\uparrow$)} & 44.6 {\small \color{deepgreen}(18.1$\uparrow$)} & 45.5 {\small \color{deepgreen}(19.1$\uparrow$)} & 42.6 {\small \color{deepgreen}(15.1$\uparrow$)} \\
&&& AA & 52.4 {\small \color{deepgreen}(42.1$\uparrow$)} & 78.6 {\small \color{deepgreen}(32.4$\uparrow$)} & 27.4 {\small \color{deepgreen}(17.4$\uparrow$)} & 13.1 {\small \color{deepgreen}(10.1$\uparrow$)} & 70.0 {\small \color{deepgreen}(51.5$\uparrow$)} & 8.6 {\small \color{deepgreen}(8.0$\uparrow$)} & 53.0 {\small \color{deepgreen}(46.3$\uparrow$)} & 40.5 {\small \color{deepgreen}(31.0$\uparrow$)} & 34.7 {\small \color{deepgreen}(33.0$\uparrow$)} & 44.9 {\small \color{deepgreen}(37.5$\uparrow$)} & 44.8 {\small \color{deepgreen}(37.6$\uparrow$)} & 42.5 {\small \color{deepgreen}(31.4$\uparrow$)} \\
\midrule
\multirow{12}{*}{\rotatebox{90}{\textbf{V-L Independent}}} & 
\multirow{6}{*}{APT-VLI} &
\multirow{3}{*}{ViT-B/16} &
PGD & 24.3 & 65.3 & 18.9 & 10.0 & 43.6 & 3.1 & 14.3 & 23.6 & 10.5 & 18.2 & 17.4 & 22.7 \\
&&& DI & 35.1 & 67.7 & 19.3 & 10.6 & 49.8 & 4.3 & 23.6 & 27.7 & 16.2 & 26.4 & 21.5 & 27.5 \\
&&& AA & 17.2 & 57.1 & 14.4 & 8.2 & 35.3 & 1.5 & 8.9 & 16.6 & 5.1 & 11.9 & 12.5 & 17.2 \\
\cline{3-16}
&&\multirow{3}{*}{ViT-B/32} &
PGD & 21.2 & 63.2 & 15.8 & 10.4 & 37.6 & 1.5 & 16.1 & 20.2 & 8.3 & 16.9 & 17.6 & 20.8 \\
&&& DI & 28.9 & 69.5 & 20.3 & 10.9 & 47.0 & 2.8 & 24.8 & 24.3 & 14.8 & 23.8 & 23.6 & 26.4 \\
&&& AA & 9.7 & 45.6 & 10.6 & 6.8 & 17.3 & 0.4 & 6.0 & 8.7 & 2.0 & 6.8 & 6.9 & 11.0 \\
\cline{2-16}
& 
\multirow{6}{*}{\textbf{TAPT-VLI}} &
\multirow{3}{*}{ViT-B/16} &
PGD & 50.0 {\small \color{deepgreen}(25.7$\uparrow$)} & 79.0 {\small \color{deepgreen}(13.7$\uparrow$)} & 32.4 {\small \color{deepgreen}(13.5$\uparrow$)} & 36.2 {\small \color{deepgreen}(26.2$\uparrow$)} & 67.5 {\small \color{deepgreen}(51.4$\uparrow$)} & 13.1 {\small \color{deepgreen}(10.0$\uparrow$)} & 65.7 {\small \color{deepgreen}(51.4$\uparrow$)} & 49.8 {\small \color{deepgreen}(26.2$\uparrow$)} & 39.6 {\small \color{deepgreen}(29.1$\uparrow$)} & 48.3 {\small \color{deepgreen}(30.1$\uparrow$)} & 47.5 {\small \color{deepgreen}(30.1$\uparrow$)} & 48.1 {\small \color{deepgreen}(25.4$\uparrow$)} \\
&&& DI & 53.8 {\small \color{deepgreen}(18.7$\uparrow$)} & 80.5 {\small \color{deepgreen}(12.8$\uparrow$)} & 32.3 {\small \color{deepgreen}(13.0$\uparrow$)} & 39.6 {\small \color{deepgreen}(29.0$\uparrow$)} & 69.4 {\small \color{deepgreen}(19.6$\uparrow$)} & 13.3 {\small \color{deepgreen}(9.0$\uparrow$)} & 70.6 {\small \color{deepgreen}(47.0$\uparrow$)} & 51.1 {\small \color{deepgreen}(23.4$\uparrow$)} & 42.5 {\small \color{deepgreen}(26.3$\uparrow$)} & 50.3 {\small \color{deepgreen}(23.9$\uparrow$)} & 48.2 {\small \color{deepgreen}(26.7$\uparrow$)} & 50.1 {\small \color{deepgreen}(22.6$\uparrow$)} \\
&&& AA & 55.1 {\small \color{deepgreen}(37.9$\uparrow$)} & 80.3 {\small \color{deepgreen}(23.2$\uparrow$)} & 35.7 {\small \color{deepgreen}(21.3$\uparrow$)} & 44.4 {\small \color{deepgreen}(36.2$\uparrow$)} & 70.2 {\small \color{deepgreen}(34.9$\uparrow$)} & 16.0 {\small \color{deepgreen}(14.5$\uparrow$)} & 76.2 {\small \color{deepgreen}(67.3$\uparrow$)} & 55.1 {\small \color{deepgreen}(38.5$\uparrow$)} & 50.5 {\small \color{deepgreen}(45.4$\uparrow$)} & 53.6 {\small \color{deepgreen}(41.7$\uparrow$)} & 54.5 {\small \color{deepgreen}(42.0$\uparrow$)} & 53.8 {\small \color{deepgreen}(36.6$\uparrow$)} \\
\cline{3-16}
&&\multirow{3}{*}{ViT-B/32} &
PGD & 48.2 {\small \color{deepgreen}(27.0$\uparrow$)} & 82.1 {\small \color{deepgreen}(18.9$\uparrow$)} & 30.7 {\small \color{deepgreen}(14.9$\uparrow$)} & 31.6 {\small \color{deepgreen}(21.2$\uparrow$)} & 68.1 {\small \color{deepgreen}(30.5$\uparrow$)} & 4.5 {\small \color{deepgreen}(3.0$\uparrow$)} & 64.7 {\small \color{deepgreen}(48.6$\uparrow$)} & 44.6 {\small \color{deepgreen}(24.4$\uparrow$)} & 41.3 {\small \color{deepgreen}(33.0$\uparrow$)} & 47.6 {\small \color{deepgreen}(30.7$\uparrow$)} & 49.1 {\small \color{deepgreen}(31.5$\uparrow$)} & 46.6 {\small \color{deepgreen}(25.8$\uparrow$)} \\
&&& DI & 50.0 {\small \color{deepgreen}(21.1$\uparrow$)} & 82.7 {\small \color{deepgreen}(13.2$\uparrow$)} & 31.5 {\small \color{deepgreen}(11.2$\uparrow$)} & 32.7 {\small \color{deepgreen}(21.8$\uparrow$)} & 69.4 {\small \color{deepgreen}(22.4$\uparrow$)} & 4.4 {\small \color{deepgreen}(1.6$\uparrow$)} & 65.6 {\small \color{deepgreen}(40.8$\uparrow$)} & 45.1 {\small \color{deepgreen}(20.8$\uparrow$)} & 42.8 {\small \color{deepgreen}(28.0$\uparrow$)} & 49.1 {\small \color{deepgreen}(25.3$\uparrow$)} & 49.9 {\small \color{deepgreen}(26.3$\uparrow$)} & 47.6 {\small \color{deepgreen}(21.2$\uparrow$)} \\
&&& AA & 49.7 {\small \color{deepgreen}(40.0$\uparrow$)} & 80.6 {\small \color{deepgreen}(35.0$\uparrow$)} & 33.3 {\small \color{deepgreen}(22.7$\uparrow$)} & 38.1 {\small \color{deepgreen}(31.3$\uparrow$)} & 68.4 {\small \color{deepgreen}(51.1$\uparrow$)} & 5.0 {\small \color{deepgreen}(4.6$\uparrow$)} & 67.5 {\small \color{deepgreen}(61.5$\uparrow$)} & 47.1 {\small \color{deepgreen}(38.4$\uparrow$)} & 48.1 {\small \color{deepgreen}(46.1$\uparrow$)} & 50.3 {\small \color{deepgreen}(43.5$\uparrow$)} & 50.4 {\small \color{deepgreen}(43.5$\uparrow$)} & 49.0 {\small \color{deepgreen}(38.0$\uparrow$)} \\
\bottomrule
\end{tabular}}}
\caption{Zero-shot adversarial robustness (\%) of different defense methods from ImageNet to downstream datasets, evaluated against PGD~\cite{madry2017towards}, DI~\cite{xie2019improving}, and AutoAttack (AA)~\cite{croce2020reliable} under perturbation budget $\epsilon=1/255$. The baseline APT methods (APT-V, APT-VLJ, and APT-VLI) were tuned on ImageNet under a 16-shot setting and then assessed on the other 10 datasets. The green upward arrows ($\color{deepgreen}\uparrow$) highlight the performance improvement of our TAPT over the baselines.}
\label{tab:1}
\end{table*}

\begin{table}[htbp]
\centering
\resizebox{1.0\linewidth}{!}{
\setlength{\tabcolsep}{1.2mm}{
\begin{tabular}{lccccccccccccccccc}
\toprule
& \textbf{Source} & \multicolumn{11}{c}{\textbf{Target}} \\ 
\cmidrule(lr){2-2} \cmidrule(lr){3-13}
& \rotatebox{90}{ImageNet} 
& \rotatebox{90}{Caltech101} 
& \rotatebox{90}{DTD} 
& \rotatebox{90}{EuroSAT} 
& \rotatebox{90}{Pets} 
& \rotatebox{90}{Aircraft} 
& \rotatebox{90}{Food101} 
& \rotatebox{90}{Flowers} 
& \rotatebox{90}{Cars} 
& \rotatebox{90}{SUN397} 
& \rotatebox{90}{UCF101} 
& \rotatebox{90}{\textbf{Avg.}}\\
 \midrule
   APT-V      & 60.6 & 89.7 & 36.9 & 26.7 & 84.8 & 16.8 & 63.2 & 55.7 & 52.1 & 58.2 & 54.4 & 54.5\\
   \textbf{TAPT-V}   & \textbf{66.9} & \textbf{92.8} & \textbf{44.6} & \textbf{41.0} & \textbf{88.5} & \textbf{23.8} & \textbf{85.7} & \textbf{67.4} & \textbf{65.7} & \textbf{62.9} & \textbf{65.5} & \textbf{64.1}\\
 \midrule
   APT-VLJ   & 64.0 & 88.5 & 34.8 & 16.0 & \textbf{81.1} & \textbf{8.1} & 62.6 & \textbf{53.3} & 47.0 & 53.0 & 52.8 & 51.0\\
   \textbf{TAPT-VLJ} & \textbf{66.5} & \textbf{88.5} & \textbf{36.7} & \textbf{16.7} & 80.6 & 7.4 & \textbf{64.6} & 50.4 & \textbf{48.7} & \textbf{54.7} & \textbf{55.4} & \textbf{51.8}\\
 \midrule
   APT-VLI    & 63.8 & 88.6 & 34.1 & 17.2 & 80.7 & 11.6 & 61.1 & 51.8 & 45.9 & 53.5 & 52.7 & 51.0\\
   \textbf{TAPT-VLI} & \textbf{65.8} & \textbf{90.4} & \textbf{39.2} & \textbf{48.3} & \textbf{82.5} & \textbf{16.3} & \textbf{85.8} & \textbf{60.3} & \textbf{62.5} & \textbf{60.0} & \textbf{64.4} & \textbf{61.4}\\
\bottomrule
\end{tabular}}}
\caption{Zero-shot clean accuracy (\%) of different defense methods from ImageNet to downstream datasets, including APT baselines (APT-V, APT-VLJ, and APT-VLI) and our TAPT. The backbone is ViT-B/16. The best results are \textbf{boldfaced}.}
\label{tab:2}
\end{table}

\subsection{Experimental Setup}
\noindent\textbf{Datasets and Models}
We experiment on 11 benchmark datasets (ImageNet test set and 10 other zero-shot test datasets): ImageNet~\cite{russakovsky2015imagenet}, Caltech101~\cite{fei2004learning}, DTD~\cite{cimpoi2014describing}, EuroSAT~\cite{helber2019eurosat}, Pets~\cite{parkhi2012cats}, Aircraft~\cite{maji2013fine}, Food101~\cite{bossard2014food}, Flowers~\cite{nilsback2008automated}, Cars~\cite{krause20133d}, SUN397~\cite{xiao2010sun}, and UCF101~\cite{soomro2012ucf101}. Our experiments focus on the CLIP model, specifically utilizing the ViT-B/16 and ViT-B/32 architectures. Following standard CLIP usage, we used hand-crafted prompts as textual inputs. For example, the prompt \texttt{"a photo of a <class>, a type of pet"} was applied for the Pets dataset. A summary of these datasets and their corresponding hand-crafted prompts are provided in the Appendix.

\begin{figure*}[htbp]
    \centering
    \includegraphics[width=1\linewidth]{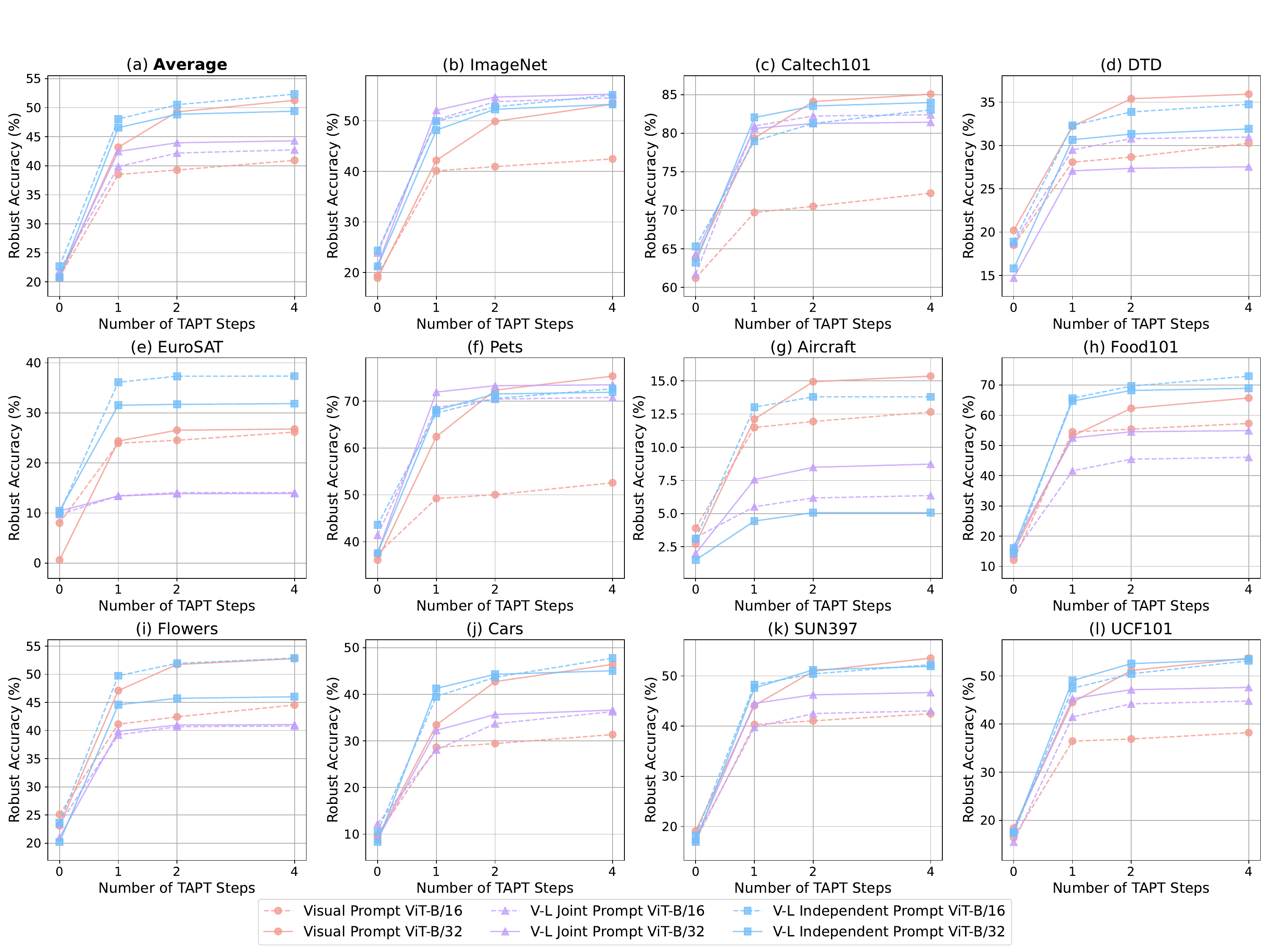}
   \caption{Adversarial robustness (\%) of our TAPT method under different test-time adaptation steps (i.e., \{0, 1, 2, 4\}). The results are reported against the PGD-100 attack on ViT-B/16 and ViT-B/32 architectures.}
    \label{fig4}
    \vspace{-0.1in}
\end{figure*}

\vspace{0.1cm}  
\noindent\textbf{Attack Configuration}  
We evaluate the zero-shot adversarial robustness of CLIP against both white-box and black-box attacks. Specifically, we employ PGD-100~\cite{madry2017towards} for white-box attacks, DI~\cite{xie2019improving} for black-box attacks, and the more powerful AutoAttack~\cite{croce2020reliable}. The hyperparameters for PGD-100 and DI were configured based on the TorchAttacks library~\cite{kim2020torchattacks}. Consistent with~\cite{mao2023understanding}, we use perturbation budgets of $\epsilon = 1/255$, $2/255$, and $4/255$ for both attacks.

\vspace{0.1cm}  
\noindent\textbf{Defense Configuration}  
For existing APT methods, we use the original configurations of APT methods~\cite{zhang2023adversarial, li2024one} and generate adversarial examples using PGD-2 attack with step size $\alpha = 1/255$. We then develop more robust versions of APT methods with different prompt designs, including 1) adversarial visual prompt tuning (APT-V), 2) adversarial V-L joint prompt tuning (APT-VLJ), and 3) adversarial V-L independent prompt tuning (APT-VLI).

\vspace{0.1cm}  
\noindent\textbf{Implementation Details}  
For our TAPT method, we initialize the defensive prompt using APT on ImageNet, training for 100 epochs with a batch size of 32 and a learning rate of 0.035. To generate augmented views for test-time fine-tuning, we create 63 variations of each test sample using random resized crops and horizontal flips, resulting in 64 images per sample, including the original. From these 64 images, we select the top 10\% most confident predictions (with the lowest entropy) and compute the average entropy of their predicted probabilities. For adversarial-clean alignment, we pre-compute embedding statistics from the public dataset (ImageNet) using APT and standard PT, respectively. We then optimize the defensive prompts by minimizing a combined loss of multi-view entropy and adversarial-clean alignment using the AdamW optimizer, with a learning rate of $5 \times 10^{-4}$ and an adversarial-clean scale factor of $\alpha = 0.5$, on a single NVIDIA A100 GPU.

\subsection{Main Results}
\noindent\textbf{Zero-Shot Adversarial Robustness}  
We compare our TAPT method with existing APT methods across three prompt designs: visual-only (V), V-L joint (VLJ), and V-L independent (VLI). Table~\ref{tab:1} presents the zero-shot adversarial robustness results under PGD-100, DI, and AutoAttack attacks, with the `Vanilla' showing results without any defense. 
It is clear that both white-box and black-box attacks can drastically reduce accuracy (nearly 0\%) in the absence of defenses, using only imperceptible noise with $\epsilon=1/255$. AutoAttack emerges as the most effective attack, achieving a nearly 100\% attack success rate on average across all datasets. 
For defense, our TAPT method consistently achieves the best average performance under AutoAttack across all prompt designs (visual-only, V-L joint, and V-L independent) and ViT architectures (ViT-B/16 and ViT-B/32). 
With ViT-B/16, TAPT enhances robustness by 32.6\% (visual-only), 25.9\% (V-L joint), and 36.6\% (V-L independent), respectively. Similar improvements are observed with ViT-B/32, where TAPT yields gains of 34.1\%, 31.4\%, and 38.0\% for the respective prompt designs. TAPT also demonstrates superior performance against both PGD-100 and DI attacks.

Furthermore, TAPT with the V-L independent prompt design achieves the highest average zero-shot adversarial robustness, surpassing both visual-only and V-L joint prompt designs. To summarize, our experiments reveal that: (1) incorporating textual prompts generally improves zero-shot adversarial robustness across various datasets; (2) the V-L joint prompt enhances robustness on the source domain (ImageNet) more effectively than other prompt designs; and (3) the V-L independent prompt design is more effective in enhancing zero-shot adversarial robustness than the V-L joint prompt design, likely due to the challenges in optimizing the interplay between visual and textual prompts.

\vspace{0.1cm}
\noindent\textbf{Zero-Shot Clean Accuracy}  
Table~\ref{tab:2} presents the zero-shot clean accuracy of different defense methods from ImageNet to downstream datasets, comparing APT baselines (APT-V, APT-VLJ, and APT-VLI) with our TAPT method on the ViT-B/16 architecture. 
Note that the APT baselines were trained on ImageNet and subsequently tested on 10 downstream datasets, while for our TAPT, only the robust statistics were computed based on ImageNet.
The zero-shot clean accuracy should be compared on all 10 datasets. 
As shown, TAPT consistently achieves superior clean performance across all 11 datasets, demonstrating significantly stronger generalization capabilities. While a slight performance decrease is observed with TAPT's V-L joint prompt design on Pets, Aircraft, and Flowers, it remains competitive. Overall, TAPT outperforms APT across all three prompt designs (visual-only, V-L joint, and V-L independent), with average improvements of 9.6\%, 0.8\%, and 10.4\%, respectively. These results underscore TAPT's ability to improve adversarial robustness without compromising too much clean accuracy.

\subsection{Ablation Studies}
\noindent\textbf{Number of TAPT Steps}  
We first examine the effect of the number of test-time adaptation steps on the robust accuracy of TAPT. Figure~\ref{fig4} shows TAPT's robustness performance with varying steps (0, 1, 2, and 4) under the PGD-100 attack. Notably, $\text{step}=0$ represents the baseline performance without TAPT, reflecting the adversarial robustness achieved solely through APT. A clear trend emerges, showing that robust accuracy increases with additional TAPT steps across most datasets and prompt designs. While robustness gains generally stabilize after a few TAPT steps, the improvement from 0 to 1 step is consistently substantial, highlighting the effectiveness of even a single adaptation step. 
However, datasets vary in sensitivity to the number of TAPT steps; for example, performance stabilizes quickly on datasets like EuroSAT, whereas for datasets such as ImageNet and DTD, further improvements are observed with additional steps. The figure also illustrates that TAPT’s benefits are consistent across different prompt designs. 
We further analyzed the per-sample time overhead of TAPT, finding that the additional inference time per image is 0.095s (visual-only), 0.166s (V-L joint), and 0.165s (V-L independent), respectively. This demonstrates that TAPT not only consistently enhances robust accuracy but also maintains a relatively low time cost.

\begin{figure}[htbp]
    \centering
    \includegraphics[width=1\linewidth]{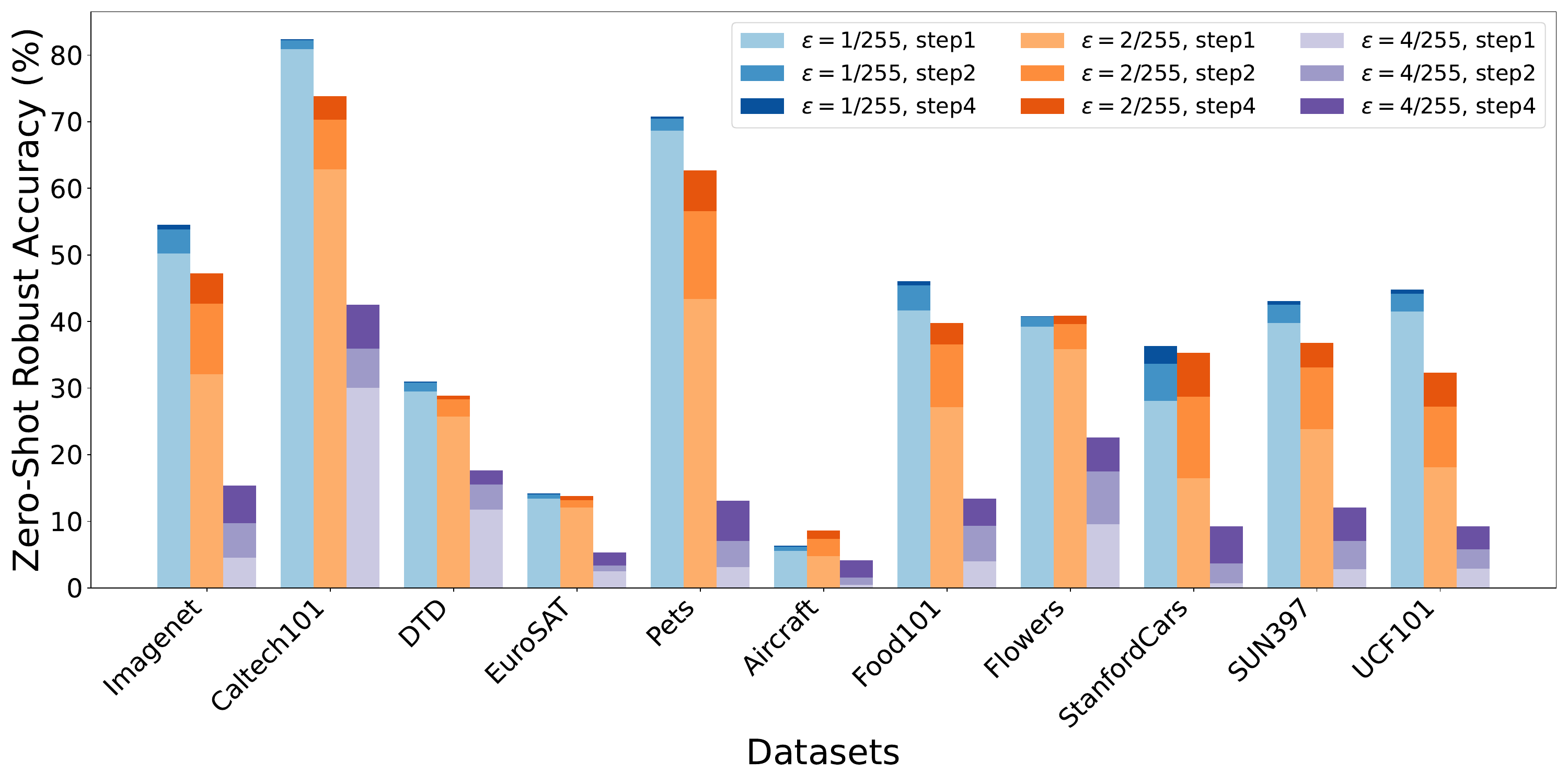}
    \caption{Zero-shot adversarial robustness (y-axis) ofs TAPT under varying perturbation budgets $\epsilon$ (1/255, 2/255, and 4/255) and TAPT steps (1, 2, and 4).}
    \label{fig5}
\end{figure}

\vspace{0.1cm}
\noindent\textbf{Different Perturbation Budgets}  
We further assess TAPT's robustness under varying attack strengths, defined by the perturbation budget $\epsilon$. Figure~\ref{fig5} displays zero-shot adversarial robustness across 11 datasets with $\epsilon$ values of 1/255, 2/255, and 4/255, and TAPT steps of 1, 2, and 4. As expected, robust accuracy decreases as $\epsilon$ increases, indicating stronger attacks. Nonetheless, TAPT consistently enhances robust accuracy across all datasets and $\epsilon$ values. 

\noindent\textbf{TAPT Reset Intervals}  
Our TAPT method resets the prompt to its initial state before processing each test sample, ensuring that each test sample is handled independently. However, we also explored alternative strategies with varied reset intervals. As shown in Table~\ref{tab:3}, we experimented with reset intervals ranging from 1 to 32, as well as a strategy with no reset during the entire inference process (i.e., ``reset=all"). Our findings indicate that continuous prompt adaptation without frequent resets can further improve robust accuracy, suggesting that accumulating information across multiple inputs can be beneficial. However, this continuous strategy introduces a vulnerability to potential poisoning attacks. To mitigate this risk and ensure reliable test-time defense, we recommend the per-sample reset strategy (``reset=1"), which prioritizes robustness against potential attacks over marginal gains from continued TAPT.

\begin{table}[htbp]
    \centering
    \resizebox{0.9\linewidth}{!}{
    \setlength{\tabcolsep}{2.0mm}{
    \begin{tabular}{lcc}
    \toprule
    \textbf{Reset Interval} & \textbf{ImageNet} & \textbf{10 Zero-Shot Datasets Avg.} \\
    \midrule
    reset=1   & 49.92          & 47.85          \\
    reset=2   & 50.20          & \textbf{47.91} \\
    reset=4   & 50.69          & \textbf{47.91} \\
    reset=8   & 51.14          & 47.53          \\
    reset=16  & 51.49          & 46.39          \\
    reset=32  & \textbf{51.62} & 44.50          \\
    reset=all & 0.48           & 3.79           \\
    \bottomrule
    \end{tabular}}}
    \caption{Zero-shot adversarial robustness (\%) of TAPT with varying reset intervals on ImageNet and 10 other zero-shot datasets. ``reset = $N$'' means the prompt is reset after every $N$ test samples.}
    \label{tab:3}
\end{table}

\section{Limitation}
\label{sec:limit}
As a test-time defense method, TAPT has certain limitations that warrant further research. Our method primarily addresses attacks in the image modality by aligning adversarial image embeddings with pre-computed public data statistics on a public dataset. Future work could explore additional modality alignment and acceleration techniques to facilitate TAPT’s deployment in industrial applications. Moreover, our current focus is limited to image recognition tasks. Extending TAPT to a broader range of tasks, such as visual reasoning and visual question answering in advanced models like GPT-4V~\cite{openai2023gpt} and Gemini~\cite{team2023gemini}, represents a promising direction for future research.
\section{Conclusion}
\label{sec:conclusion}
In this paper, we introduced a novel test-time defense method, \emph{Test-Time Adversarial Prompt Tuning (TAPT)}, to enhance the inference robustness of pre-trained VLMs, such as CLIP. TAPT tunes defensive bimodal (textual and visual) prompts for each test sample by leveraging multi-view entropy minimization and adversarial-clean alignment, effectively safeguarding CLIP’s zero-shot inference. It also utilizes pre-computed statistics from a public dataset (ImageNet) to defend a wide range of downstream tasks. Comprehensive evaluation across 11 benchmark datasets demonstrates that TAPT effectively enhances zero-shot adversarial robustness against both white-box and black-box attacks, while largely preserving clean accuracy. Compared to training-time adversarial prompt tuning (APT) methods, TAPT offers several advantages: (1) it is unsupervised, (2) enables sample-wise prompt adaptation, (3) delivers superior zero-shot adversarial robustness and clean accuracy, (4) is independent of downstream tasks, and (5) is lightweight. Future research could explore the scalability of test-time defense across different modalities.

{
    \small
    \bibliographystyle{ieeenat_fullname}
    \bibliography{main}
}


\end{document}